\documentclass{article}

\usepackage{arxiv}

\usepackage[utf8]{inputenc} 
\usepackage[T1]{fontenc}    
\usepackage{hyperref}       
\usepackage{url}            
\usepackage{booktabs}       
\usepackage{amsfonts}       
\usepackage{nicefrac}       
\usepackage{microtype}      
\usepackage{lipsum}
\usepackage{tikz}
\usepackage{amsmath, amssymb}
\usepackage{xfrac}
\usepackage{mathtools}
\usepackage{algorithmic}
\usepackage{multicol}
\usepackage{listings}
\usepackage{romannum,upgreek}

\newcommand{\var}[1]
{
\text{\itshape{#1}}
}

\usepackage{scalerel}
\usetikzlibrary{svg.path}

\definecolor{orcidlogocol}{HTML}{A6CE39}
\tikzset{
  orcidlogo/.pic={
    \fill[orcidlogocol] svg{M256,128c0,70.7-57.3,128-128,128C57.3,256,0,198.7,0,128C0,57.3,57.3,0,128,0C198.7,0,256,57.3,256,128z};
    \fill[white] svg{M86.3,186.2H70.9V79.1h15.4v48.4V186.2z}
                 svg{M108.9,79.1h41.6c39.6,0,57,28.3,57,53.6c0,27.5-21.5,53.6-56.8,53.6h-41.8V79.1z M124.3,172.4h24.5c34.9,0,42.9-26.5,42.9-39.7c0-21.5-13.7-39.7-43.7-39.7h-23.7V172.4z}
                 svg{M88.7,56.8c0,5.5-4.5,10.1-10.1,10.1c-5.6,0-10.1-4.6-10.1-10.1c0-5.6,4.5-10.1,10.1-10.1C84.2,46.7,88.7,51.3,88.7,56.8z};
  }
}

\newcommand\orcidicon[1]{\href{https://orcid.org/#1}{\mbox{\scalerel*{
\begin{tikzpicture}[yscale=-1,transform shape]
\pic{orcidlogo};
\end{tikzpicture}
}{|}}}}

\newtheorem{hyp}{Hypothesis}

\definecolor{codegreen}{rgb}{0,0.5,0}
\definecolor{codegray}{rgb}{0.5,0.5,0.5}
\definecolor{codered}{rgb}{0.75, 0.3 ,0.1}
 
\lstdefinestyle{mystyle}{
    commentstyle=\color{codegreen},
    keywordstyle=\color{codered},
    numberstyle=\tiny\color{codegray},
    basicstyle=\ttfamily\footnotesize,
    breakatwhitespace=false,         
    breaklines=true,                 
    captionpos=b,                    
    keepspaces=true,                 
    numbers=left,                    
    numbersep=5pt,                  
    showspaces=false,                
    showstringspaces=false,
    showtabs=false,                  
    tabsize=2
}
 
\DeclareMathOperator*{\argmax}{argmax}

\title{Scheduling optimization of parallel linear algebra algorithms using Supervised Learning}

\author{
  Gabriel ~Laberge \\
    Department of Mathematics \\
    and Industrial Engineering\\
    Polytechnique Montr\'eal\\
    Montr\'eal, QC, Canada \\
    \texttt{gabriel.laberge@polymtl.ca}\\
  \And
  Shahrzad ~Shirzad\\
    Center for Computation and Technology \\
    Louisiana State University\\
    Baton-Rouge, LA, USA \\
    \texttt{sshirz1@lsu.edu}\\
  \And
  Patrick ~Diehl\\
    Center for Computation and Technology\\ 
    Louisiana State University\\
    Baton-Rouge, LA, USA \\
    \orcidicon{0000-0003-3922-8419}{0000-0003-3922-8419}\\
  \And
  Hartmut Kaiser\\
    Center for Computation and Technology \\
    Louisiana State University\\
    Baton-Rouge, LA, USA \\
    \texttt{hkaiser@cct.lsu.edu}
  \And
  Serge Prudhomme\\
    Department of Mathematics\\
    and Industrial Engineering\\
    Polytechnique Montr\'eal\\
    Montr\'eal, QC, Canada \\
    \texttt{serge.prudhomme@polymtl.ca}\\
  \And
  Adrian S. Lemoine\\
    Center for Computation and Technology \\
    Louisiana State University\\
    Baton-Rouge, LA, USA \\
    \texttt{aserio@cct.lsu.edu}\\
}

\begin{document}
\maketitle

\begin{abstract}
Linear algebra algorithms are used widely in a variety of domains, e.g machine learning, numerical physics and video games graphics. For all these applications, loop-level parallelism is required to achieve high performance. However, finding the optimal way to schedule the workload between threads is a non-trivial problem because it depends on the structure of the algorithm being parallelized and the hardware the executable is run on. In the realm of Asynchronous Many Task runtime systems, a key aspect of the scheduling problem is predicting the proper chunk-size, where the chunk-size is defined as the number of iterations of a for-loop assigned to a thread as one task. In this paper, we study the applications of supervised learning models to predict the chunk-size which yields maximum performance on multiple parallel linear algebra operations using the HPX backend of Blaze's linear algebra library. More precisely, we generate our training and tests sets by measuring performance of the application with different chunk-sizes for multiple linear algebra operations; vector-addition, matrix-vector-multiplication, matrix-matrix addition and matrix-matrix-multiplication. 
We compare the use of logistic regression, neural networks and decision trees with a newly developed decision tree based model in order to predict the optimal value for chunk-size. Our results show that classical decision trees and our custom decision tree model are able to forecast a chunk-size which results in good performance for the linear algebra operations.
\end{abstract}


\section{Introduction}
\label{sec:intro}

The efficient scheduling of tasks within parallel loops is a problem that must be solved in every instance of loop-level parallelism, no matter the application. 
A key aspect to efficient execution of these loops is to manage the amount of work contained in each task. 
This amount of work per task is known as its chunk-size. The creation of tasks induces overheads, however, the more tasks created the more work can be done concurrently. The challenge for application developers is to schedule in a way such that the overheads introduced by new tasks are amortized by the performance gains made by the increase in parallelism. 

The simplest loop scheduling method is static scheduling.
This method divides the iterations evenly and statically among all the processors, either as a consecutive block or in a round-robin manner~\cite{liu1994safe}. Since all the assignments happen at compile time or before execution of the application, this method imposes no run-time scheduling overhead. Several factors including interprocessor communication, cache misses, and page faults can lead to different execution times for different iterations, leading to load imbalance among the processors~\cite{philip1995increasing}.
Meanwhile, dynamic scheduling methods postpone the assignment to run-time, which tends to improve load balancing, at the cost of higher scheduling overhead. Some of the dynamic scheduling methods include: Pure Self-scheduling, Chunk Self-scheduling, Guided Self-scheduling~\cite{polychronopoulos1987guided}, Factoring~\cite{hummel1992factoring} and Trapezoid Self-scheduling~\cite{tzen1993trapezoid,liu1994safe}.

In this paper we focus solely on dynamic scheduling methods. These approaches, require the fine tuning the chunk-size as chunk-sizes which are too small will cause scheduling and execution overheads and chunk-sizes too large will not induce enough parallelism to achieve maximum performance. This ``sweet spot'' is application-dependent because it varies with the structure of the task and the hardware used. In particular, we are interested in parallelizing the linear algebra algorithms in the Blaze C++ library using the HPX backend.
Blaze is a linear algebra library that uses smart expression templates in order to mix readable mathematical syntax and high performance~\cite{iglberger2012expression}. HPX is a C++ standard library used for parallelism and concurrency~\cite{kaiser2014hpx,heller2017hpx}. 

The objective of this research is to generate a machine learning based scheduler that is able to take compile-time and run-time information of any linear algebra operation and predict the optimal chunk-size at run-time with minimal overhead to make the predictions. This scheduler should be the least-intrusive within the Blaze source code and should not disturb the normal work-flow of the user. We investigate if a supervised learning model can be used to make such a scheduler.

Our work is an extension of~\cite{khatami2017hpx} where machine learning was used in tandem with HPX for the first time. In this research, the authors developed a ClangTool to extract compile-time information by analyzing the Abstract Syntax Tree of any algorithm being parallelized. For a large set of parallel algorithms, they extracted information like the number of operations per iteration, number of floating point operations per iteration, the number of \lstinline|if| statements, the deepest loop level and number of function calls per iteration, number of iterations and number of threads. The training and test sets where generated by using the ClangTool on various algorithms. The features extracted were used as the inputs of a logistic regression algorithm to predict the best chunk-size at run-time. The novelty of the ClangTool is that it can work for a large set of algorithms since it extracts low-level information from the abstract syntax tree directly. However, it would not apply to our linear algebra operations because we use dynamic matrices. Since the size of the matrices are unknown at compile-time, values like number of operations per iteration cannot be extracted by a compiler.

The major contributions of the paper are the following:
\begin{itemize}
\item We extend the existing range of machine learning applications in HPC by using supervised learning to optimize the performance of basic parallel linear algebra operations.
\item We compare existing approaches to compute the optimal chunk-size for a given operation and conclude that we can directly predict the chunk-size via a classification method that provides efficient and accurate predictions. 
\item We develop a new decision-tree classification model that is customized to more efficiently estimate the optimal chunk-size. Such a model is, in theory, not restricted to the linear algebra operations described in this paper. It could be used to solve any new problem where the objective is to predict the value of a certain parameter that yields optimal performance for a given application.
\end{itemize}

The paper is organized as follows:
In Section~\ref{sec:rleated:work}, we review the literature on contributions related to our research work. Machine learning terminology, definition of the parallelization framework, and mathematical formulation of the scheduling as an optimization problem are discussed in Section~\ref{sec:metho}. Section~\ref{sec:learning} describes our methodology to solve the optimization problem via supervised learning and our experimental results are shown in Section~\ref{sec:results}. Finally, we conclude with Section~\ref{sec:conclusion}.

\section{Related Contributions}
\label{sec:rleated:work}
Most of the previous work done in this field focuses on predicting the performance of parallel algorithms, which is different from what we are trying to accomplish. However, modeling performance can be used to find the optimal way to schedule a loop and it is linked to our research. Previous works mainly focus on three types of models: analytic, trace-based, and empirical models~\cite{malakar2018benchmarking}. 

Analytic models~\cite{blagojevic2008modeling,kerbyson2001predictive,valiant1990bridging}, while providing an arithmetic formula to represent the execution time of an application, require a deep understanding of the application. These models cannot be generalized to other types of domains and architectures~\cite{lee2007methods,sun2017automated,pllana2007performance}.

Traced-based models use the traces collected through instrumentation in order to predict  performance. Instrumentation refers to the code that is added to the program source code as a way to capture run-time information. These models, in contrast to analytic models, do not rely on an expert's knowledge of the application. However, they do add some overhead at run-time, require a large storage space to save the traces and are hard to analyze~\cite{sun2017automated}. Moreover, for any new application, one must first run a portion of the code before producing performance predictions.

In empirical modeling, the results are obtained by running an application on multiple machines with a set of user-specified features to build a model that will predict performance for new feature values~\cite{malakar2018benchmarking}. This type of modeling includes machine learning-based approaches. The line between empirical models and trace based models can be blur at times as traces can be seen as the features of an empirical model. The two methods differ, however, by the fact that trace-based models add extra instructions to the program, which is not necessarily the case with empirical models.

It is our observation that three types of features are commonly used in literature for empirical models, e.g.\ non-deterministic features, deterministic application-free features, and deterministic application-specific features. Non-deterministic features are values that keep changing when re-running the same program. These include various performance counters, like cache hits-misses, CPU idle rate etc. Deterministic application-free features are values that do not change when running the same program and can be measured on different kinds of applications. For example the authors in~\cite{khatami2017hpx} used deterministic application-free features extracted by compiler. Values like number of operations per iteration, deepest loop level, number of threads, etc.\ are not specific to a certain task. Deterministic application-specific features are parameters that can only be computed for a certain class of algorithms and therefore cannot be used for other algorithms. For example, matrix sizes could be used to characterize a linear algebra operations but it cannot be used to characterize a binary tree search algorithm. This terminology, despite being non-official, will be used throughout the paper when talking about features for machine learning models in order to be more specific.

In~\cite{ipek2005approach}, the authors use neural networks to predict the performance for SMG2000 applications, a parallel multi-grid solver for linear systems~\cite{falgout2002hypre}, on two different architectures. They use a fully connected neural network to predict the performance, defining workloads per processor and processor topology as the deterministic application-free features of the model. Since they consider the absolute mean error as their loss function, they use stratification to replicate samples with lower values of performance by a factor that is proportional to their performance as a way to ensure that no sample is masked. They also apply bagging techniques to decrease the variance in the model. As they increase the size of the training set to 5K points, they reach an error rate of 4.9\%. 

The trace-based model~\cite{sun2017automated} analyzes the abstract syntax tree of a code and collects data by inserting special code for instrumentation in four different situations: assignments, branches, loops, and MPI communications. These features have the advantage of being application-free so they can be used to characterize different operations: Graph500, GalaxSee, and SMG2000.
The authors then use different machine learning methods, e.g.\ random forests, support vector machine, and ridge regression to build a prediction model from the collected data. By applying a two feature reduction processes, they were able to decrease the amount of additional overhead and storage requirements. Their results indicated that the random forests method was the most useful, because of the lower impact of categorical features on it, which is helpful in the general cases where one has no knowledge about the type of features~\cite{sun2017automated}.

In~\cite{malakar2018benchmarking}, the authors investigate a set of machine learning techniques, including deep neural networks, support vector machine, decision tree, random forest, and k-nearest neighbors to predict the execution time of four different applications. They use deterministic application-specific features for each of their applications. For example, in the case of the miniMD molecular dynamics application, the number of processes and the number of atoms were considered input features, while for miniAMR, an application for studying adaptive mesh refinement, the number of processes and block sizes in the $x$, $y$, and $z$ directions, were used as the input features. 
While achieving promising results, especially for deep neural networks, bagging, and boosting methods, the authors suggest utilizing transfer learning through deep neural networks to predict performance on other platforms.

Despite concentrating on GPUs, Liu et al. ~\cite{liu2018runtime} proposed a lightweight machine learning based performance model to choose the number of threads to use for the parallelization of the training of a neural network (NN). They chose to use non-deterministic features collected by hardware counters, namely, the number of CPU cycles, the numbers of cache misses, the accesses for the last cache level, and the number of level 1 cache hits.
Since each step of the training of a NN has the same computational and memory patterns, the authors used the first steps to extract the performance counters and fed them to the performance model. The output of the model then guided the choice for the number of threads that would be used throughout the rest of the NN training steps. 
They took two different approaches to build their model. 
In the first one, they tried ten different regression models including random forest, and in the second one, they used a hill climbing algorithm to choose the number of threads. In addition of being hardware independent, the hill climbing algorithm achieves a much higher accuracy compared to the best performing regression model. The authors suspected that machine learning models were poorly performing because of inaccuracies in the hardware counters.

In this paper, we propose
using machine learning to directly predict the optimal chunk-size to achieve the best performance instead of predicting the execution time. Also, we do not attempt to find the optimal number of cores to run an application on like in ~\cite{liu2018runtime}. In our research, it is assumed that the user is working on a given number of cores and simply want to find the optimal way to share the workload between these cores. These are examples of what differentiates our approach from the ones discussed above.

\section{Mathematical formulation}
\label{sec:metho}
\subsection{Supervised learning basics}
\label{sec:metho:SL}
In supervised learning, the objective is to predict a variable $y \in \mathcal{Y}$ whose value is unknown given some other known variable $\Vec{x} \in \mathcal{X}$. This is equivalent to finding a function $f:\mathcal{X} \rightarrow \mathcal{Y}$, which will be referred from now on to as the model. The input of the model is called the feature vector and the output is called the target. When the set $\mathcal{Y}$ is finite, the model $f$ is 
considered a classification function and when $\mathcal{Y} = \mathbb{R}$, the model is 
considered a regression. The output of the model $\hat{y}=f(\Vec{x})$ is usually different from the actual value of $y$, so we define the notion of error or ``loss'' between the actual and predicted values by using a loss function $\mathcal{L}(y, \hat{y})$, which is application-dependent. The loss can be computed on multiple instances of feature vectors and targets contained within a set $\mathcal{S} = \{(\Vec{x}^{(i)}, y^{(i)})\}_{i=1}^N$ (note that the superscript $i$ will always be used to represent an instance within a set). This leads to the notion of the cost function $\mathcal{J}$ on the set $\mathcal{S}$:
\begin{equation}
    \mathcal{J} \big(f,
    \mathcal{S} \big) = \frac{1}{N} \sum_{i=1}^N \mathcal{L} \big(y^{(i)}, f(\Vec{x}^{(i)})\big).
    \label{eq:cost}
\end{equation}
Typically, the machine learning model is selected by minimizing the cost function on a so-called training set. Once a model $f$ has been selected, it can be evaluated on new input values $\vec{x}_{\var{new}}$ that were not included in the training set. To assess how the model performs on these new values, the cost function must be evaluated on a test set containing previously unused inputs. A model that performs well on the test set is said to generalize. The notion of generalization is critical to estimate how the model will perform in real applications. More details on supervised learning can be found in~\cite{hastie2005elements}.

\subsection{Parallelization Framework}
\label{sec:metho:parallel}
\begin{figure}[tbp]
\centering
\resizebox{8cm}{3cm}{
\begin{tikzpicture}
\draw[fill=black!10!white] (0, 0) rectangle (1, 1);
\draw[fill=black!10!white] (1, 1) rectangle (2, 2);
\draw[fill=black!10!white] (0, 1) rectangle (1, 2);
\draw[fill=black!10!white] (1, 0) rectangle (2, 1);

\draw[fill=black!10!white] (4.1, -0.1) rectangle (5.1, 0.9);
\draw[fill=black!10!white] (2.8, 1.1) rectangle (3.8, 2.1);
\draw[fill=black!10!white] (4.1, 1.1) rectangle (5.1, 2.1);
\draw[fill=black!10!white] (2.8, -0.1) rectangle (3.8, 0.9);

\draw[fill=black!10!white] (7.1, -0.1) rectangle (8.1, 0.9);
\draw[fill=black!10!white] (5.8, 1.1) rectangle (6.8, 2.1);
\draw[fill=black!10!white] (7.1, 1.1) rectangle (8.1, 2.1);
\draw[fill=black!10!white] (5.8, -0.1) rectangle (6.8, 0.9);

\draw (6.95, 1.6) ellipse (1.6cm and 0.72cm);
\node at (7, 2.7) {chunk-size=2};
\draw [<-, line width=0.5mm] (10,1.5) -- (8.54,1.5);
\draw [] (3.3, 1.62) circle (0.72cm);
\node at (3.3, 2.7) {block-size $\in \mathbb{N}^2$};
\node at (11,1.5) {Thread 1};
\node at (11,0.5) {Thread 2};
\node at (11,-0.5) {...};
\draw[->, line width = 0.5mm] (0,-1) -- (11,-1);
\node at (5.5,-1.5) {Time};
\node at (1,2.5) {$C$};
\end{tikzpicture}
}
\caption{Distribution of the computation of the elements of the matrix C among threads. First, the matrix C is segmented into sub-matrices of size ``block-size''. These sub-matrices are then assembled into chunks of size ``chunk-size'', which are dynamically assigned to threads.}
\label{fig:parallel}
\end{figure}
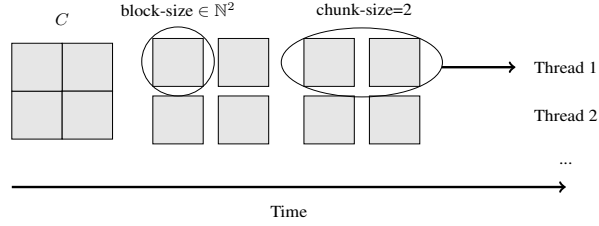

In the Blaze library, every linear algebra expression, like $A + B$, is accessible as an expression template that is a light weight object. More complex expressions are stored in expression template trees that can be traversed at compile time via template meta-programming. The elements of an expression are only evaluated when the assignment operator is called, i.e.\ $C = \text{ExpressionTemplate}(A, B)$. During an assignment, the elements of the expression template are evaluated and the results are stored in $C$. This step can be parallelized since each component of $C$ is independently computed. Figure~\ref{fig:parallel} shows how work is distributed among threads and introduces the two critical parameters involved: block-size and chunk-size. In this work, we demonstrate the use of machine learning to simplify the task of selecting the chunk sizes, given a specified block-size.

\subsection{Mathematical Formulation of the Scheduling Problem}
\label{sec:metho:math}
We discuss here the mathematical formulation that will be used throughout the paper as well as a more formal statement of the scheduling problem. Many variables are implicated in the problem to solve so we introduce the following notation and domains:

\begin{multicols}{2}
\begin{itemize}
    \item Matrix size (Vector size): $\var{ms} \in \var{MS} \subset \mathbb{N}$
    \item Number of threads: $N_{\var{thr}} \in \{2, 4, 6. ..., 12, 14, 16\}$
    \item Operation type: $\var{ot} \in \var{OT}$
    \item Architecture: $\var{ar} \in \var{AT}$
    \item Chunk-size: $\var{cs} \in \var{CS} \subset \mathbb{N}$
    \item Block-size: $\var{b} \in \var{B} \subset \mathbb{N}^2$
    \item Unaccounted processes: $\epsilon$
    \item[\vspace{\fill}]
\end{itemize}
\end{multicols}

$\var{OT}$ represents the domain of all possible linear algebra operations and $\var{AT}$ represents the set of all possible architectures. These sets are too complex to be written explicitly. The variable $\epsilon$ describes all processes in the computer that we are unaware of. Our ignorance of this variable introduces a non-deterministic behavior to the system. It is assumed that there exists a function $\text{Per}$ that takes all these variables as inputs and outputs the performance in MFlop/s:
\begin{align}
\text{MFlop/s} = \text{Per}(\var{ms}, N_{\var{thr}}, \var{ot}, \var{ar}, \var{b}, \var{cs})\text{.}
\end{align}
Note that the dependence of \text{Per} on $\epsilon$ is not explicitly written so its influence will be perceived as noise. By defining
$\Vec{x} := (\var{ms}, N_{\var{thr}}, \var{ot}, \var{ar})$ as an abstract feature vector, we can write the following optimization problem:
\begin{align}
    \text{Given}\ \Vec{x},\ \text{find}\ (\var{cs}^*, \var{b}^*) = \argmax_{(\var{cs}, \var{b}) \in \var{CS} \times B} \,\,  \text{Per}(\Vec{x},\var{b},\var{cs})\text{.}
    \label{eq:optim}
\end{align}
The values $\var{cs}^*$ and $\var{b}^*$ will be referred to as the optimal chunk-size and block-size, respectively. We use machine learning terminology for $\Vec{x}$ because solving Eq.~(\ref{eq:optim}) is analogous to finding a model $f:\mathcal{X} \rightarrow \var{CS} \times \var{B}$ with input $\Vec{x}$. We say that the feature vector is abstract because the components $\var{ar}$ and $\var{ot}$ do not take numerical values but represent concepts. For example, $\var{ot}$ can stand for \textit{matrix-matrix multiplication}, \textit{vector-vector addition}, etc. This abstract feature vector is useful when doing equations with a general notation, but all of its abstract components will need to be transformed into numbers when it comes to generate the training and test sets. We will refer to this as \textit{concretizing} the abstract feature vector. The deterministic application-free features, deterministic application-specific features, and non-deterministic features described earlier can be used to \textit{concretize} the abstract components.

One way to solve Eq.~(\ref{eq:optim}) would be to use Zero-Order Optimization. These methods do an intelligent search in the space $\var{CS} \times B$ to find maximum performance. Such methods are very expensive because the function $\text{Per}$ must be called multiple times to find a maximum. Performance is costly to compute so we cannot consider such methods to find the optimal parameters at run-time.

The previous HPX backend scheduler solved Eq.~(\ref{eq:optim}) by fixing $\var{cs} = 1$ and choosing the block-size so the matrix would be divided into $N_{\var{thr}}$ sub-matrices of similar sizes. For example, the block-size chosen for a $\var{ms}\times \var{ms}$ matrix operation on 4 threads would be $(\lceil\sfrac{\var{ms}}{2}\rceil, \lceil\sfrac{\var{ms}}{2}\rceil)$ which is the specific blocking scenario illustrated in Figure~\ref{fig:parallel}. With this blocking, it is assumed that $\var{b}$ depends only on $\var{ms}$ and $N_{\var{thr}}$. However, since $\var{b}$ is linked to memory, it should depend as well on $\var{at}$ and $\var{ot}$ since different architectures may have different cache sizes and different operations may access memory differently. 

The next section describes the proposed approach to solve this optimization problem by using supervised learning models.

\section{Methodology}
\label{sec:learning}
\subsection{Benchmarks and Resources}
\label{sec:learning:setup}
All experiments were performed on a node consisting
of two Intel\textsuperscript{\textregistered} 
Xeon\textsuperscript{\textregistered}
CPU E5-2450 clocked at 2.10GHZ,
with 16 cores, hyperthreading disabled, and 48 GB DDR4 memory. The CPUs had L1 caches of 32 kBytes and non-shared L2 caches of 256 kBytes. Blaze v3.4 was compiled with the dependencies OpenBlas v0.3.2, and HPX v1.3.0~\cite{hartmut_kaiser_2019_3189323}. HPX was compiled using Boost 1.68.0 and Hwloc 1.11.0 as its dependencies.  The compiler used for Blaze and HPX was Clang 6.0.1. All benchmarks that were used in this paper are listed in Table~\ref{tab:allben}. Mat, Vec, D, T refer to matrix, vector, dense and transpose, respectively. For example, DVecDVecAdd means an addition of two dense vectors. The default alignment of matrices in Blaze is row-major so transposed matrices are column-major.

\begin{table}[tbp]
\label{tab:allben}
\begin{center}
\begin{tabular}{cc}
\toprule
Daxpy        & TDMatTDMatAdd  \\ \hline
DVecDVecAdd  & DMatScalarMult \\ \hline
DMatDVecMult & DMatDMatMult   \\ \hline
DMatDMatAdd  & DMatTDMatMult  \\ \hline
DMatTDMatAdd & TDMatTDMatMult \\ \hline
\multicolumn{2}{c}{TDMatDMatAdd}          \\ \bottomrule
\end{tabular}
\end{center}
\caption{Benchmarks investigated in this paper. Mat, Vec, D, T refer to matrix, vector, dense and transpose, respectively.  }
\end{table}

\subsection{Estimating the Best Block-size}
\label{sec:learning:bestblock}
Solving Eq.~(\ref{eq:optim}) for both $\var{cs}^*$ and $\var{b}^*$ is rather intensive because of the amount of configurations in $\var{CS}\times \var{B}$. To simplify the problem, we decided to first solve for $\var{b}^*$ and then for $\var{cs}^*$. To find the optimal block-size, we consider the following hypothesis:

\begin{hyp}
$\var{b}^* = \var{b}^*(\var{ar}, \var{ot})$, i.e.\
the best block-size depends only on architecture and operation types.
\label{hyp:hyp1}
\end{hyp}

Our reasoning is that finding the best block size is a hardware related problem. The optimal block-size must ensure that all the data sent to a CPU must fit within its cache. Since different benchmarks will access the memory differently, it is also safe to assume that the best block-size will depend on the operation type. The approach used to compute $\var{b}^*$ was to fix $\var{cs}$ and start with $\var{b}$ containing $\var{b}_1 \times \var{b}_2 = 4096$ elements and change its aspect ratio $\sfrac{\var{b}_1}{\var{b}_2}$ to see how it impacted performance. For some benchmarks where these values did not give good performance, other values were found empirically. This approach is somewhat heuristic, but it is not the focal point of this paper. Indeed, our main goal is to predict the best chunk-size when given a certain blocking. For that reason, the block-sizes to be used on the benchmarks can be chosen sub-optimal.

\subsection{Estimating the Best Chunk-size}
\label{sec:learning:bestchunk}
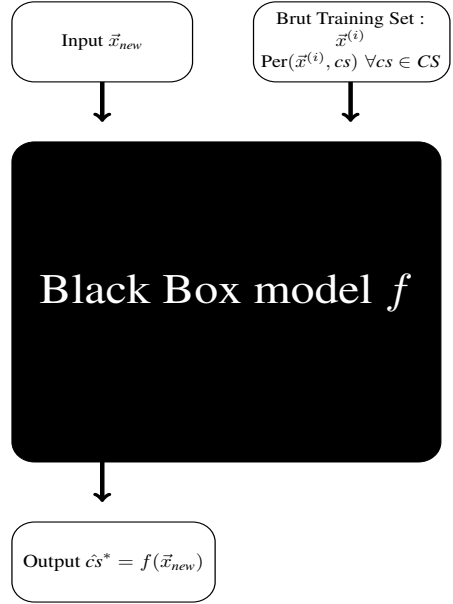
\begin{figure}[tbp]
\centering
\resizebox{6cm}{8cm}{
\begin{tikzpicture}

\tikzstyle{myedgestyle} = [ultra thick]

\draw[fill=black, rounded corners=5mm]  (-4.5,3.5) rectangle (5,-4.5);

\draw [rounded corners=5mm, thick] (-4.5,7) rectangle (-0.5,5);
\draw [rounded corners=5mm, thick] (0.85,7) rectangle (5.2,5);
\draw [rounded corners=5mm, thick] (-4.5,-6) rectangle (0,-8);

\node[scale=1.25] at (-2.5,6) {Input $\vec{x}_{\var{new}}$};
\node[scale=1.25] at (3,6) {\begin{tabular}{c}
    Brut Training Set : \\
    $\Vec{x}^{(i)}$  \\
    $\text{Per}(\Vec{x}^{(i)}, \var{cs}) \, \, \forall \var{cs} \in \var{CS}$ \\
\end{tabular}};
\node[scale=1.3] at (-2.25,-7) {Output $\hat{\var{cs}}^* = f(\vec{x}_{\var{new}})$};
\node[scale=3, text=white] at (0.25,-0.25) {Black Box model $f$};

\draw [->, line width=1mm](-2.5,5) -- (-2.5,4);
\draw [->, line width=1mm](3,5) -- (3,4);
\draw [->, line width=1mm](-2.5,-4.5) -- (-2.5,-5.5);

\end{tikzpicture}}
\caption{A black box classification model. This model is trained by looking at examples $i$. Each example consists of a feature vector and performances for all possible chunk-sizes in $\var{CS}$. After training, the black box model is able to predict the optimal chunk-size for a new value of the feature vector.}
\label{fig:blackbox}
\end{figure}

Once the approximation of the best block-size is computed for any $\var{ot}$ and $\var{ar}$, we can redefine the abstract feature vector $\Vec{x} := (\var{ms}, N_{\var{thr}}, \var{ot}, \var{ar}, \var{b}^*)$. The best chunk-size can be calculated by solving the problem:
\begin{align}
    \text{Given}\ \Vec{x},\ \text{find}\ \var{cs}^*=\argmax_{\var{cs} \in \var{CS}} \,\, \text{Per}(\Vec{x}, \var{cs})\text{.}
    \label{eq:optim_cs}
\end{align}
We have used supervised learning models to find the solutions of Eq.~(\ref{eq:optim_cs}) for any arbitrary value of the feature vector $\Vec{x}$. The advantage of machine learning methods is that they can output predictions pretty fast. Therefore, the selection of the optimal chunk-size can be done seamlessly in Blaze. The most time-consuming part of the machine learning methodology is generating data and training the model. However, these models do not need to be trained often. Our goal is to find a black-box model fed with a brut training set $\{(\Vec{x}^{(i)},\ \text{Per}(\Vec{x}^{(i)},\var{cs})\ \forall \var{cs} \in \var{CS})\}_{i=1}^N$, which outputs an approximation of the best chunk-size for any new example $\hat{\var{cs}}^*=f(\vec{x}_{\var{new}})$ (where the hat symbols stands for approximation). The reason we call this set the brut training set will become clear later on. Such a model is illustrated in Figure~\ref{fig:blackbox}. We refer to this model as a black box model because there exist multiple ways to implement them as discussed below. However, all black box models share the same input, output, and brut-training set. They only differ by how they are implemented in the ``inside'' which justifies the terminology. We describe two types of these black box models that were seen in literature and attempt to give them formal names that will be used throughout the paper as a way to distinguish both.

\subsubsection{Pre-Training Optimization Model (PreTO)}
This model is the same as the one used in~\cite{khatami2017hpx} to compute the optimal chunk-size, see Figure~\ref{fig:PreTO}. The goal is to solve Eq.~(\ref{eq:optim_cs}) for each example in the brut training set to generate pairs of values $\mathcal{S}=\{(\Vec{x}^{(i)}, \var{cs}^{*(i)})\}_{i=1}^N$ to be used for the training of the classification model. The target of the machine learning model is then $\var{cs}^*$, the value we want to predict.  Classification is chosen because the output set  $\var{CS}\subset \mathbb{N}$ is a finite set. The output of the black-box model $f(\vec{x}_{\var{new}})$ is directly the output of the classification model within the black box.

\begin{figure}
\centering
\begin{minipage}[t]{\dimexpr.5\textwidth-1em}
\centering

\resizebox{7cm}{9cm}{
\begin{tikzpicture}

\tikzstyle{myedgestyle} = [ultra thick]

\draw[fill=black!0!white, rounded corners=5mm, line width = 1mm]  (-4.5,3.5) rectangle (5,-4.5);

\draw [->, line width=1mm](-2.5,5) -- (-2.5,4);
\draw [->, line width=1mm](3,5) -- (3,4);
\draw [->, line width=1mm](-2.5,-3.5) node (v1) {} -- (-2.5,-5.5);

\draw [rounded corners=5mm, thick] (-4.5,7) rectangle (-0.5,5);
\draw [rounded corners=5mm, thick] (0.85,7) rectangle (5.2,5);
\draw [rounded corners=5mm, thick] (-4.5,-5.6) rectangle (-0.1,-7.4);

\draw[fill=black!0!white, rounded corners=5mm, thick] (0.4,3) rectangle (4.8,0.5);
\draw[fill=black!0!white, rounded corners=5mm, thick] (1,-0.5) rectangle (4.5,-2);
\draw[fill=black!0!white, rounded corners=5mm, thick]  (-4,-1) rectangle (-0.5,-3.5);

\draw[->, draw=black,line width=1mm] (-2.5,3) -- (-2.5,-0.5);
\draw[->, draw=black,line width=1mm] (2.5,0.5) -- (2.5,-0.5);
\draw[->, draw=black,line width=1mm] (2.5,-2) -- (2.5,-2.75) -- (0,-2.75);

\node[scale=1.3] at (-2.5,6) {Input $\vec{x}_{\var{new}}$};
\node[scale=1.25] at (3,6) {\begin{tabular}{c}
    Brut Training Set : \\
    $\Vec{x}^{(i)}$  \\
    $\text{Per}(\Vec{x}^{(i)}, \var{cs}) \, \, \forall \var{cs} \in \var{CS}$ \\
\end{tabular}};
\node[scale=1.3] at (-2.3,-6.5) {Output $\hat{\var{cs}}^* = f(\vec{x}_{\var{new}})$};

\node[align=center, scale=1.5] at (-2.25,-2.28) {Classification \\
Model $f$};
\node[align=center, scale=1] at (2.6,1.65) {Optimization: \\
$\var{cs}^{*(i)} =\underset{ \var{cs} \in \var{CS}}{\mathrm{argmax}} \, \text{Per}(\Vec{x}^{(i)},\var{cs})$ };
\node[align=center, scale=1] at (2.7,-1.25) {Training Set \\ $\mathcal{S}=(\Vec{x}^{(i)}, \var{cs}^{*(i)})$};

\node[text=black, align=center, scale=1] at (2,-3.55) {Feature Scaling \\ Training};
\node[text=black, align=center, scale=1, rotate=-90] at (-2,1.5) {Feature Scaling};

\end{tikzpicture}
}
\caption{Inside of the Pre-Training Optimization Model. It is called so because the optimization of performance on the training set is done before training the machine learning model. The classification model learns to associate a feature vector $\Vec{x}$ with the optimal chunk size correspondent. The output of the model $f(\vec{x}_{\var{new}})$ is the direct output of the classifier}
\label{fig:PreTO}

\end{minipage}\hfill
\begin{minipage}[t]{\dimexpr.5\textwidth-1em}
\centering

\resizebox{7cm}{9cm}{
\begin{tikzpicture}

\tikzstyle{myedgestyle} = [ultra thick]

\draw[fill=white, rounded corners=5mm, line width = 1mm]  (-4.5,3.5) rectangle (5,-4.5);
\draw[fill=black!0!white, rounded corners=5mm]  (-4,-3) rectangle (-0.5,-4);
\draw[fill=black!0!white, rounded corners=5mm] (-4,3) rectangle (4.5,1.5);
\draw[fill=black!0!white, rounded corners=5mm] (1,0.5) rectangle (4.5,-0.5);
\draw[fill=black!0!white, rounded corners=5mm]  (-4,0) rectangle (-0.5,-2);

\draw [->, line width=1mm](-2.5,5) -- (-2.5,4);
\draw [->, line width=1mm](3,5) -- (3,4);
\draw [->, line width=1mm](-2.5,-4) node (v1) {} -- (-2.5,-5.5);

\draw [rounded corners=5mm, thick] (-4.5,7) rectangle (-0.5,5);
\draw [rounded corners=5mm, thick] (0.85,7) rectangle (5.2,5);
\draw [rounded corners=5mm, thick] (-4.5,-5.6) rectangle (-0.1,-7.4);

\draw[->, draw=black,line width=1mm] (-2.5,1.5) -- (-2.5,0);
\draw[->, draw=black,line width=1mm] (2.5,1.5) -- (2.5,0.5);
\draw[->, draw=black,line width=1mm] (2.5,-0.5) -- (2.5,-1.5) -- (0,-1.5);
\draw[->, draw=black,line width=1mm] (-2.5,-2) -- (-2.5, -3);


\node[scale=1.3] at (-2.5,6) {Input $\vec{x}_{\var{new}}$};
\node[scale=1.25] at (3,6) {\begin{tabular}{c}
    Brut Training Set : \\
    $\Vec{x}^{(i)}$  \\
    $\text{Per}(\Vec{x}^{(i)}, \var{cs}) \, \, \forall \var{cs} \in \var{CS}$ \\
\end{tabular}};
\node[scale=1.3] at (-2.3,-6.5) {Output $\hat{\var{cs}}^* = f(\vec{x}_{\var{new}})$};

\node[align=center, scale=1.5] at (-2.25,-1) {Regression \\
Model\\
$\widehat{\text{Per}}(\Vec{z})$};

\node[align=center, scale=1] at (0.45,2.25) {Feature Augmentation and Scaling: \\
$\Vec{z} = (\Vec{x}, \var{cs})$ };

\node[align=center, scale=1] at (2.7,0) {Training Set \\ $\mathcal{S}'=(\Vec{z}^{(i)}, \text{Per}(\vec{z}^{(i)}))$};

\node[text=black, align=center, scale=1] at (1.75,-2) {Training};
\node[scale=1] at (-2.25,-3.5) {$\hat{\var{cs}}^* =\underset{ \var{cs} \in \var{CS}}{\mathrm{argmax}} \, \widehat{\text{Per}}(\Vec{z})$};
\draw[draw=black, line width=1mm];
\end{tikzpicture}
}
\caption{Inside of the Post-Training Optimization Model. Here the optimization of performance is done after training the machine learning model. The training data is used to model a regression of performance and then for any new example, this regression function is maximized with respect to chunk-size to get the prediction.}
\label{fig:PosTO}

\end{minipage}
\end{figure}

Classical machine learning classification algorithms try to maximize the accuracy on the training set:
\begin{align}
\text{Acc} \big(f, \mathcal{S} \big) = \frac{1}{N} \sum_{i=1}^{N} \mathbb{I} \big(\var{cs}^{*(i)} = f(\Vec{x}^{(i)})\big) 
\label{eq:accuracy}
\end{align}
where the loss function, $\mathbb{I}(.)$, is an indicator function that returns one if the input statement is true and zero otherwise. Note that, in practice, classification algorithms rarely maximize the accuracy directly because it is non-differentiable. Many classification algorithm solve equivalent optimization problems instead. For example, logistic regressions and neural networks use log-likelihood as their loss-function. For decision trees, the accuracy is replaced by the information gain~\cite{hastie2005elements}. All these methods are indirect ways of maximizing the
accuracy so we can only consider the accuracy without loss of generality. The advantage of the preTO is that it implies very little overhead as the only step required to compute a prediction is to feed the input to the classification model inside the box. The problem with the PreTO model is that, sometimes, different chunk-sizes yield similar performance and the classification model is oblivious to that fact. Since classification models maximize accuracy, they will try to classify each chunk size on the training set correctly even if, in some cases, a miss-classification does not affect performance. This can be seen as wasted effort by the model.

\subsubsection{Post-Training Optimisation Model (PosTO)}
This model is similar to what was used in~\cite{liu2018runtime}, see Figure~\ref{fig:PosTO}. It consists of modeling $\text{Per}$ via a regression called $\widehat{\text{Per}}$ and then, instead of solving Eq.~(\ref{eq:optim_cs}), we find
\begin{align}
    \hat{\var{cs}}^* = \argmax_{\var{cs} \in \var{CS}} \,\, \widehat{\text{Per}}(\Vec{x}, \var{cs})\text{.}
    \label{eq:optim_cshat}
\end{align}

This shows how our work and all publications about performance modeling discussed in Section ~\ref{sec:rleated:work} are related. Solving Eq.~(\ref{eq:optim_cshat}) can be done with a line-search at run-time. Since we want our machine learning model to predict the performance, we need to do a feature augmentation procedure on our data. This procedure includes $\var{cs}$ among the features of the performance model instead of treating it as a target.
We can write this as defining a new feature vector $\Vec{z} := (\Vec{x}, \var{cs})$. The procedure is illustrated in Figure ~\ref{fig:augmentation}.

\begin{figure}[tbp]
\centering
\resizebox{10cm}{4cm}{
\begin{tikzpicture}

\draw (-4.4,2.6) -- (-4.4,0.8);
\draw (-4.8,2.2) -- (0,2.2);
\node at (-4.6,2.4) {$i$};
\node at (-4.6,1.8) {1};
\node at (-4.6,1.2) {2};
\node at (-4,2.4) {$x^{(i)}$};
\node at (-1,2.4) {$\text{Per}(x^{(i)}, 2)$};
\node at (-2.8,2.4) {$\text{Per}(x^{(i)}, 1)$};

\node at (-4,1.2) {1.5};
\node at (-4,1.8) {2.3};
\node at (-2.8,1.8) {1000};
\node at (-1,1.8) {2100};
\node at (-2.8,1.2) {2500};
\node at (-1,1.2) {1200};
\node at (-2.4,3.2) {Before Feature Augmentation};

\draw (1,2.2) -- (4,2.2);
\draw (1.4,2.6) -- (1.4,-0.4);

\node at (2.6,3.2) {After Feature Augmentation};
\node at (1.2,2.4) {$i$};
\node at (2,2.4) {$z^{(i)}$};
\node at (1.2,1.8) {1};
\node at (1.2,1.2) {2};
\node at (1.2,0.6) {3};
\node at (1.2,0) {4};
\node at (3.4,2.4) {$\text{Per}(z^{(i)})$};

\node at (2,1.8) {(2.3, 1)};
\node at (2,1.2) {(2.3, 2)};
\node at (2,0.6) {(1.5, 1)};
\node at (2,0) {(1.5, 2)};
\node at (3.4,1.8) {1000};
\node at (3.4,1.2) {2100};
\node at (3.4,0.6) {2500};
\node at (3.4,0) {1200};

\end{tikzpicture}
}
\caption{Hypothetical feature augmentation procedure with $\var{CS}=\{1, 2\}$ and only one feature $x$. Feature augmentation is a way of reorganizing the brut training set by including the chunk size as a feature to generate a new feature vector $z$. This has the effect of enlarging the number of examples that are part of the training set.}
\label{fig:augmentation}
\end{figure}

 After the procedure the number of examples in the training set becomes $N'=N \times |\var{CS}|$ (where $|\var{CS}|$ is the cardinality of set $\var{CS}$). The formal description of the new training set is
 $\mathcal{S}' = \{\,\big(\Vec{z}^{(i)}, \text{Per}(\Vec{z}^{(i)})\big)\,\}_{i=1}^{N'}$. Classical regression models minimize the mean-squared-error (MSE) cost function on the training set:

\begin{align}
\text{MSE}\big(\widehat{\text{Per}},\mathcal{S}' \big) = \frac{1}{N'} \sum_{i=1}^{N'} \big(\widehat{\text{Per}}(\Vec{z}^{(i)}) -\text{Per}(\Vec{z}^{(i)})\big)^2\text{.}
\label{eq:MSE}
\end{align}

The advantage of the PosTO model compared to the PreTO is that the machine learning model will take performance into account during training. The disadvantage is that there is more overhead at run-time because of the need to maximize $\widehat{\text{Per}}$ with respect to $\var{cs}$.

It is important not to confuse the black box models (PosTO and PreTO) with the machine learning models used inside them, particularly in the case of PosTO models because the outputs are different. Moreover, one must not confuse the brut training set and the training sets $\mathcal{S}$ and $\mathcal{S}'$ which are used to train the classifier and regression of performance.
The two models described earlier have the same issue; the machine learning models inside the black box have an ambiguous loss function. 
For example, in the PreTO model, the classification model could have a bad accuracy but $\text{Per}\big(\Vec{x}, f(\Vec{x})\big)$ could still be high since multiple chunk-sizes may yield good performance. Moreover, if a PreTO model has a better accuracy than another one, it does not imply that the chunk-sizes predicted by the first model will yield a higher performance than the second one. 
This is because some miss-classifications can induce a larger loss in performance. This ambiguity is only removed when $100\%$ accuracy is reached, but then, the model may be way more complex than necessary.

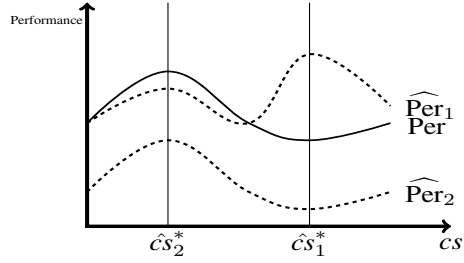
\begin{figure}[t]
\centering
\resizebox{6.3cm}{3.5cm}{
\begin{tikzpicture}

\draw[<->,line width = 1mm ](-8,-3) -- (-8,-9.5)  -- (1,-9.5);

\draw[dashed, line width=0.6mm]  plot[smooth, tension=.7] coordinates {(-8,-8.5) (-6,-7) (-4,-8.5) (-2.5,-9) (-0.5,-8.5)};
\draw[dashed, line width = 0.6mm]  plot[smooth, tension=.7] coordinates {(-8,-6.5) (-6,-5.5) (-4,-6.5) (-2.5,-4.5) (-0.5,-6)};
\draw[line width = 0.5mm]  plot[smooth, tension=.7] coordinates {(-8,-6.5) (-6,-5) (-4,-6.5) (-2.5,-7) (-0.5,-6.5)};
\node[scale=2] at (1,-10) {$\var{cs}$};
\node at (-9,-3.5) {Performance};

\node[scale=2] at (0.5,-6) {$\widehat{\text{Per}}_1$};
\node[scale=2] at (0.4,-6.6) {$\text{Per}$};
\node[scale=2] at (0.5,-8.5) {$\widehat{\text{Per}}_2$};

\draw (-6,-3) -- (-6,-9.5);
\draw (-2.5,-3) -- (-2.5,-9.5);

\node[scale=2] at (-6,-10) {$\hat{\var{cs}}^*_2$};
\node[scale=2] at (-2.5,-10) {$\hat{\var{cs}}^*_1$};

\end{tikzpicture}
}
\caption{Hypothetical case where $\Vec{x}$ is fixed. Even though the regression function $\widehat{\text{Per}}_1$ is a better fit than $\widehat{\text{Per}}_2$ in the sense of the MSE, the PosTO model that uses $\widehat{\text{Per}}_2$ as a regression of performance will predict $\hat{\var{cs}}^*_2$ which is a better choice than $\hat{\var{cs}}^*_1$. $\widehat{\text{Per}}_2$ is a better choice because it has the same ``profile'' as the actual performance, even if the values are all translated downward. Therefore, exactly predicting performance is not necessary to get good predictions of chunk-size.}
\label{fig:hypothetical}
\end{figure}

In the case of PreTO models, the MSE is also an imperfect measure for assessing the error of the black box model. In Figure~\ref{fig:hypothetical}, $\text{Per}$ is the real performance for a fixed $\Vec{x}$. The two dashed lines are hypothetical approximations of performance. One can see that $\widehat{\text{Per}}_1$ has a lower MSE than $\widehat{\text{Per}}_2$. If we solve Eq.~(\ref{eq:optim_cshat}) with both performance models, we get the predictions $\hat{\var{cs}}^*_1$ and $\hat{\var{cs}}^*_2$. However, from Figure~\ref{fig:hypothetical}, we see that
$\text{Per} \big(\Vec{x}, \hat{\var{cs}}^*_1 \big) < \text{Per} \big(\vec{x}, \hat{\var{cs}}^*_2\big) $, so the second model of performance yields a better prediction of chunk-size even though it has a larger MSE. To conclude, we cannot compare PosTO models based only on the MSE of the performance model inside them. This ambiguity is only removed when the MSE is zero, but this is surely over-fitting.

It is clear from that analysis that we need another way to assess the performance for both PreTO and PosTO models to avoid any ambiguity. Candidate measures should take the actual performances $\text{Per}\big(\Vec{x}^{(i)}, \hat{\var{cs}}^{*(i)}\big)$ into account. We introduce a new cost function that we call the Mean Sub-Optimal Performance:

\begin{align}
\text{MSOP} \big(f, \{\Vec{x}^{(i)}\}_{i=1}^N \big) = \frac{1}{N} \sum_{i=1}^{N}\frac{\text{Per}\big(\Vec{x}^{(i)}, f(\Vec{x}^{(i)})\big)}{\displaystyle \max_{\var{cs} \in \var{CS}} \text{Per}\big(\Vec{x}^{(i)}, \var{cs}\big)}\text{.}
\label{eq:MSOP}
\end{align}

The performance associated with the chunk-size prediction is divided by the optimal observed performance as a way to normalize all instances in order to ensure that benchmarks with typically better performances will not dominate every other benchmark. Also, the MSOP is restricted to take values from zero to one. The MSOP must be interpreted as \textbf{how relatively far away one is, in average, from the optimal performance that the model could have predicted}. The denominators can be computed by simply using the values in the training or test set. However, the numerators $\text{Per}\big(\Vec{x}^{(i)}, f(\Vec{x}^{(i)})\big)$, can only be easily computed if $f$ is restricted to predict from the set $\var{CS}$. In that case, performance acquired in the training or test set can be used to compute the numerators. The MSOP will be the metric used when comparing PreTO and PosTO models.

\subsection{Custom Decision Tree}
\label{sec:learning:custom}

For PreTO models, the classifier that predicts the optimal chunk-size is trained by maximizing the accuracy on the training set. The biggest limitation of accuracy is that each miss-classification is given the same impact.  Some miss-classification of chunk-size could induce bigger losses in performance than others, therefore, accuracy is not a good way to train the classification model. A reasonable loss function should be able to treat every miss-classification differently. A naive approach would be to attribute weights to each example (stratification).
Those weights could be chosen to be large for instances where one chunk-size yields a better performance than all others and small for instances where all chunk-sizes yield similar performance. 
However, for a fixed example $i$, any miss-classification still has the same impact. For this reason, using weighted accuracy is not sufficient. We decided to use the MSOP as our loss function for the classification model training process. We believe the MSOP is the right choice for the loss function because its value is only affected by a miss-classification when such miss-classifications induces a relatively large loss in performance with respect to the optimal performance. Therefore, the MSOP treats every miss-classification differently. The MSOP is non-differentiable because it involves $\text{Per}$ which one cannot differentiate. Therefore, we cannot rely on gradient-based methods for fitting. Decision trees~\cite{hastie2005elements} provide a solution to this challenge since they do not require differentiable loss-functions. Any decision tree can be described as a set of domains $D_m$ that cover the whole input space and a set of scalars $c_m$ such that:
\begin{align}
f(\Vec{x}) = \sum_{m = 1}^M c_m \mathbb{I}\big(\Vec{x} \in D_m\big)\text{.}
\label{eq:treefunc}
\end{align}
\noindent
This tree can be optimized by solving:
\begin{align}
    \argmax_{D_m, c_m, m = 1, 2, \ldots} \text{MSOP}\big(f, \{\Vec{x}^{(i)}\}_{i=1}^N\big)\text{.}
    \label{eq:tree_optim}
\end{align}
This combinatorial problem is extremely hard to solve so we attempt to find a sub-optimal solution by dividing the problem into two steps:
\begin{enumerate}
    \item Find $c_m$ given $D_m$.
    \item Find $D_m$.
\end{enumerate}

The first step is relatively straightforward and consists in solving the optimization problem:
\begin{align}
\text{Given}\ D_m ,\ c_m = \argmax_{\var{cs} \in \var{CS}}  \text{MSOP}\big(cs, \{\Vec{x}^{(i)} \in D_m\}\big)\text{.}
\label{eq:tree_optim_givenRm}
\end{align}
The MSOP is maximized by only considering the points in that domain. Since the MSOP is non-differentiable, it must be maximized via a line-search.

The second step of the optimization problem is the hardest. To find the best domains, a greedy algorithm is used. First, we start with a domain $D_1$ that covers the whole domain, we then consider a feature $j$ and a split point $s$. The domain $D_1$ is then cut into two disjoint regions $D_1(j, s) = \{\vec{x} | x_j < s \}$ and $D_2(j, s) = \{\vec{x} | x_j \geq s \}$. The best split is selected via the minimization problem:
\begin{align}
\begin{aligned}
    \min_{j,s} \sum_{k=1}^2 N_{D_k} \text{MSOP}\big(\var{c}_k, \{\Vec{x}^{(i)} \in D_k(j, s)\}\big)\text{,}
\end{aligned}
\label{eq:tree_optimRm}
\end{align}
where $N_{D_k}$, are the numbers of data points in each region. The $\var{c}_1$ and $\var{c}_2$ are computed by Eq.~(\ref{eq:tree_optimRm}) on regions $D_1$ and $D_2$. Eq. (\ref{eq:tree_optim_givenRm}) and (\ref{eq:tree_optimRm}) can be applied recursively to give a growing procedure of a custom classification tree which takes performance into account. A threshold for the MSOP in every region can be used which would stop the splits from happening in regions where the MSOP is high enough. This should result in less complex trees.
We expect the model to have less overhead on average to compute predictions than classification trees.

\subsection{Data Generation and Model Selection}
\label{seq:learning:model_selection}
The abstract feature vector is $\Vec{x} :=  (\var{ms}, N_{\var{thr}}, \var{ot}, \var{ar}, \var{b}^*)$. This vector needs to be \textit{concretized} before continuing. The compile-time variables $\var{ar}$ and $\var{ot}$ must be quantified if they are to be used in a model. In Section~\ref{sec:rleated:work}, several ways to express a feature vector in terms of numbers were suggested. In~\cite{khatami2017hpx}, deterministic application-free features were used. $\var{ot}$ was represented with characteristics of the structure of the algorithm such as number of operations, number of ``if'' statements and deepest loop level, which were measured at compile time. 
In~\cite{sun2017automated}, the authors use deterministic application-free features. They measured assignments, branches, and loops at run-time using dynamic analysis of the program. In~\cite{li2009machine}, non-deterministic features were measured. The variables $\var{ar}$, $\var{ot}$ were represented using performance counters: number of CPU cycles, number of cache misses, cache accesses for the last cache level, and number of level one cache hits. Using performance counters does not just characterize $\var{ot}$ and $\var{ar}$ but it also reveals information about the unknown processes $\epsilon$. This is because performance counters will vary based on the many unknown processes happening at the same time as the user's task. Using performance counters as features is a double edge-sword. One has reasons to think that using performance counters could improve the prediction of the model since the model will be aware of run-time imbalances between CPUs and overheads. However, if the features used to characterize $\epsilon$ are not highly correlated with performance, this will have the effect of adding unnecessary noise to the data. Moreover, in practice, every new task would need to be run twice; once to extract the performance counters and feed them to the machine learning model and another time with the predicted chunk-size for optimal performance. We don't want the user to run his linear algebra operation two times. This goes against our initial goal of not disturbing the normal work-flow of the user. For that reason, we do not use performance counters when \textit{concretizing} the feature vector. 

We also decided to remove $\var{ar}$ since we only worked on one architecture for simplicity. In theory, the proposed methodology can be extended to other architectures. To characterize the operation types, we extracted information about the complexity from Blaze's expression templates. We created a function \lstinline|getTotalMflop()| that could estimate the predicted number of floating point operations of an arbitrary expression template tree at run-time. Run-time features like $\var{ms}$ and $N_{\var{thr}}$ can be extracted at run-time with HPX's \lstinline|getnumthreads()| function and Blaze's \lstinline|matrix.cols()| member function for Matrix/Vector classes. We also changed the feature $\var{b}^*$ to $N_{\var{ite}} = \lceil \sfrac{ms}{\var{b}^*_1} \rceil \times \lceil \sfrac{ms}{\var{b}^*_2} \rceil$ which is the number of iterations in the \lstinline|for_loop|. The motivation in doing so is that the optimal chunk-size is highly correlated with $N_{\var{ite}}$. So we can define the concrete feature vector which was used when making the training and test set : $\Vec{x} := (\var{ms}, \var{Mflop}, N_{\var{thr}}, N_{\var{ite}})$. 
Note that the influence of $\var{b}^*$ is now hidden in the feature $N_{\var{ite}}$. 
Figure~\ref{tab:databen} shows all benchmarks used to generate the data-set.

\begin{table}[tbp]
\begin{center}
\begin{tabular}{cc}
\toprule
DVecDVecAdd        & DMatDVecMult \\ \hline
DMatDMatAdd & DMatScalarMult \\ \hline
DMatTDMatAdd & DMatDMatMult   \\
\bottomrule
\end{tabular}
\label{tab:databen}
\end{center}
\caption{All Benchmarks used when generating the training and test sets. Mat, Vec, D, T refer to matrix, vector, dense and transpose, respectively.  }
\end{table}

Each benchmark was run with different matrix sizes and number of threads $\{2, 4 ,6,..., 14, 16\}$ . The sets $\var{MS}$ were chosen differently for each benchmark to cover the regions of interest. The whole brut data collected consisted of 288 examples that were randomly shuffled and then split with a ratio 2:1 into a training set and a test set. To select the best black box model, we used k-fold cross-validation~\cite{hastie2005elements} on the training set and compared the average MSOP and prediction times. The value k was chosen to be 3 so the ratios would still be 2:1 when doing cross-validation. The model that performed best in average on the 3 folds, was then evaluated on the test set.
Many classical classification models and regression models can be used inside the PreTO and PosTO models. The Python library scikit-learn~\cite{scikit-learn} was used because of its high level syntax that helped to embed these models within the black-box models. For PreTO Models, both DecisionTreeClassifier and LogisticRegression were considered. For PosTO models, both MultilayerPerceptronRegression and DecisionTreeRegression were used as regressions of performance. Finally, our CustomDecisionTreeClassifier tree was implemented in Python in a PreTO model ~\cite{doi_data}.

\section{Results}
\label{sec:results}
\subsection{Best Block Size}
\label{sec:results:bestblock}
Table~\ref{tab:block} shows the values of block-size that were found empirically and the ranges of matrix sizes on which they performed well. These block-sizes were used when generating the training and test sets for the predictions of chunk-size.

\begin{table}[tbp]
\begin{center}
\begin{tabular}{c|cc}
\toprule
\textbf{Benchmark}    & \textbf{Matrix/Vector sizes} & \textbf{Selected block-size}  \\ \midrule
DVecDVecAdd & [25000, $10^6$] & 1x4096 \\ \hline
DMatDVecMult & [250, 2500] & 1x16 \\ \hline
DMatDMatAdd  &[100, 1000]  & 4x1024 \\ \hline
TDMatTDMatAdd &[100, 1000]& 1024x4 \\ \hline
DMatTDMatAdd   &[100, 1000]& 64x64  \\ \hline
TDMatDMatAdd   &[100, 1000]& 64x64  \\ \hline
DMatDMatMult & [100, $10^3$[ & 64x64 \\ \hline
DMatDMatMult & [$10^3$, $10^4$] & 256x256 \\ \hline
TDMatTDMatMult & [100, $10^3$[ & 64x64 \\ \hline
TDMatTDMatMult & [$10^3$, $10^4$] & 256x256 \\ \hline
DMatTDMatMult & [100, $10^3$[ & 64x64\\ \hline
DMatTDMatMult & [$10^3$, $10^4$] & 256x256 \\ \bottomrule
\end{tabular}
\label{tab:block}
\end{center}
\caption{Select block-size on various Blaze operations. The block-size selected are chosen to be the ones that yield the best performance for a fixed chunk-size. Matrix/Vector sizes represent the intervals of sizes where the block-size were selected}

\end{table}
We see that for matrix multiplications, the best block-size varies with the size of the matrices, which disproves hypothesis 1 (H1). We cannot explain yet why the optimal block-size increases with $\var{ms}$ for matrix multiplication. This result is hard to interpret because of the abstract nature of expression templates
and the complexity of the software stack. In Blaze, specific assignment kernels are called for each benchmark.
These Kernels can use their own blocking. Secondly, the block-size is not always representative of the number of elements indexes during an assignment. For example, with $C = A + B$ given matrices of size $\var{ms} \times \var{ms}$, a block-size of (100, 100) on $C$ will index $2 \times 100 \times 100$ elements from $A$ and $B$. However, for $C = A B$, the same block-size on $C$ will index $2 \times 100 \times \var{ms}$ elements of $A$ and $B$. For all benchmarks using square block-sizes, we found specific cases where the block-size would segment the matrix in fewer sub-matrices than there were threads, which resulted in reduced performance. It is clear that the choice of block-size must take into account the size of the Matrix/Vector and the number of threads in these cases. More research needs to be done on the blocking of these operations, however we determined this to be outside the scope of our work.

\subsection{Choice of set CS}
\label{sec:results:setCS}
Graphs of performance with respect to chunk-size were generated by running multiple benchmarks on different number of threads. Visualizing the real performance with respect to chunk-size is a useful way to design the set $\var{CS}$. It is critical to choose a good set because the black box models will be restricted to analyze examples with these chunk-sizes and will make predictions from this set. If it is too detailed, generating data will take a long time. If it is too poor, the training set will lack critical information.

\begin{figure}
\centering
\begin{minipage}[t]{\dimexpr.5\textwidth-1em}
\centering
\includegraphics[ height=5cm, width=8cm]{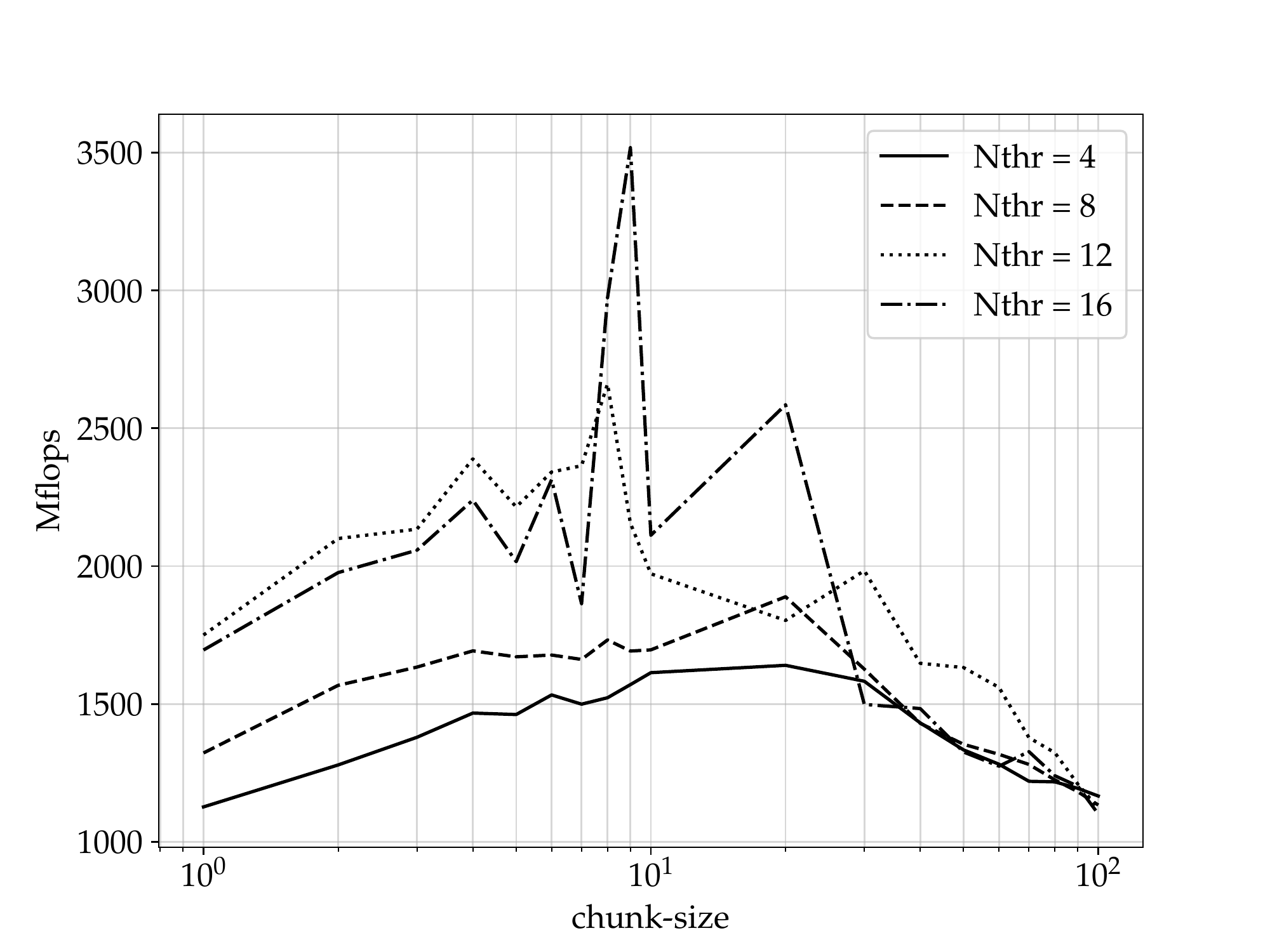}
\caption{Performance with respect to chunk-size for dense vector dense vector addition (dvecdvecadd) with ms=$10^6$ and 245 iterations  and with 4, 8, 12, 16 threads}
\label{fig:perf1}
\end{minipage}\hfill
\begin{minipage}[t]{\dimexpr.5\textwidth-1em}
\centering
\includegraphics[ height=5cm, width=8cm]{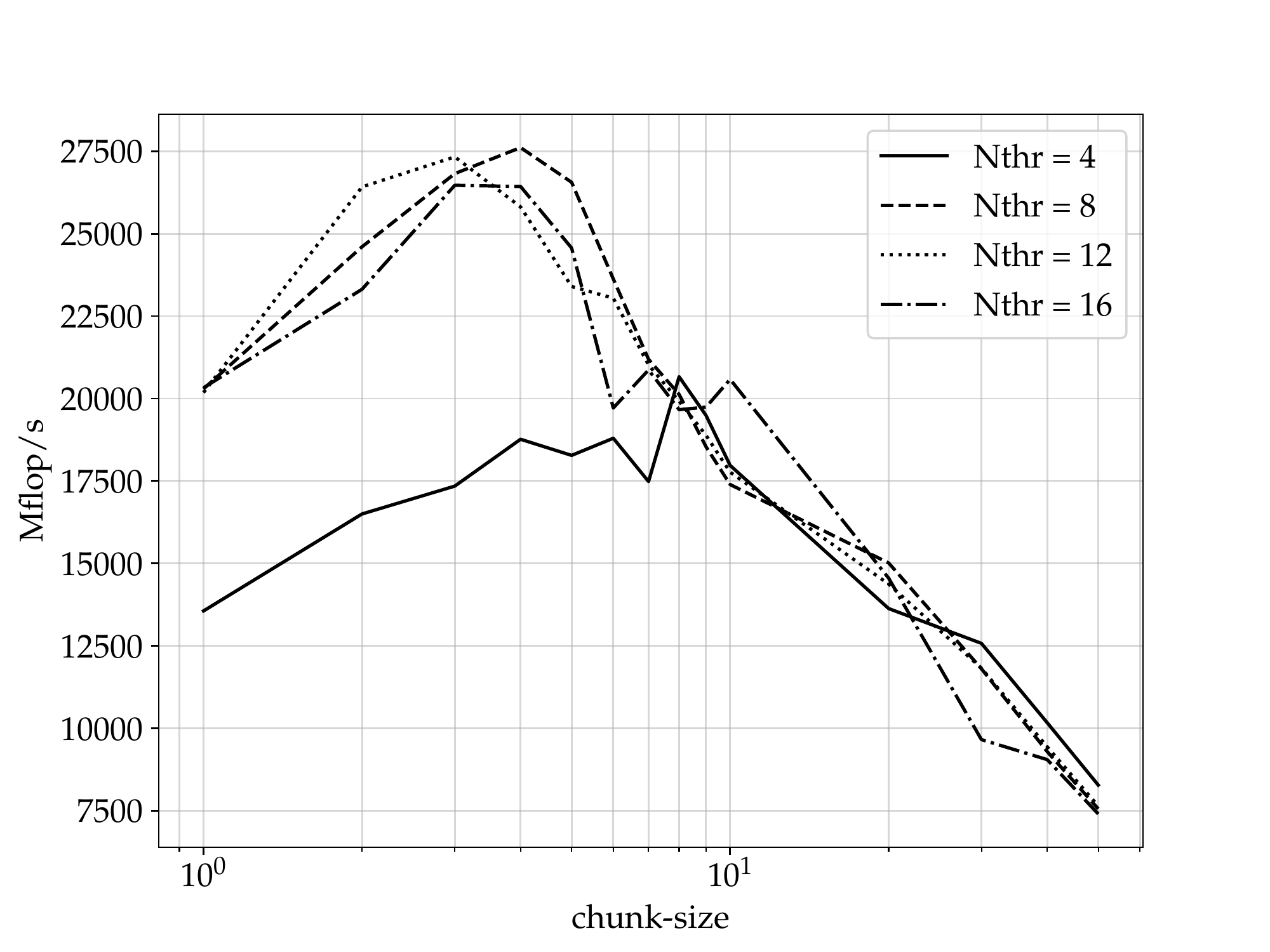}
\caption{Performance with respect to chunk-size for dense matrix dense vector multiplication (matdvecmult) with ms=100 and 63 iterations and with 4, 8, 12, 16 threads}
\label{fig:perf2}
\end{minipage}
\end{figure}

\begin{figure}
\centering
\begin{minipage}[t]{\dimexpr.5\textwidth-1em}
  \centering
\includegraphics[ height=5cm, width=8cm]{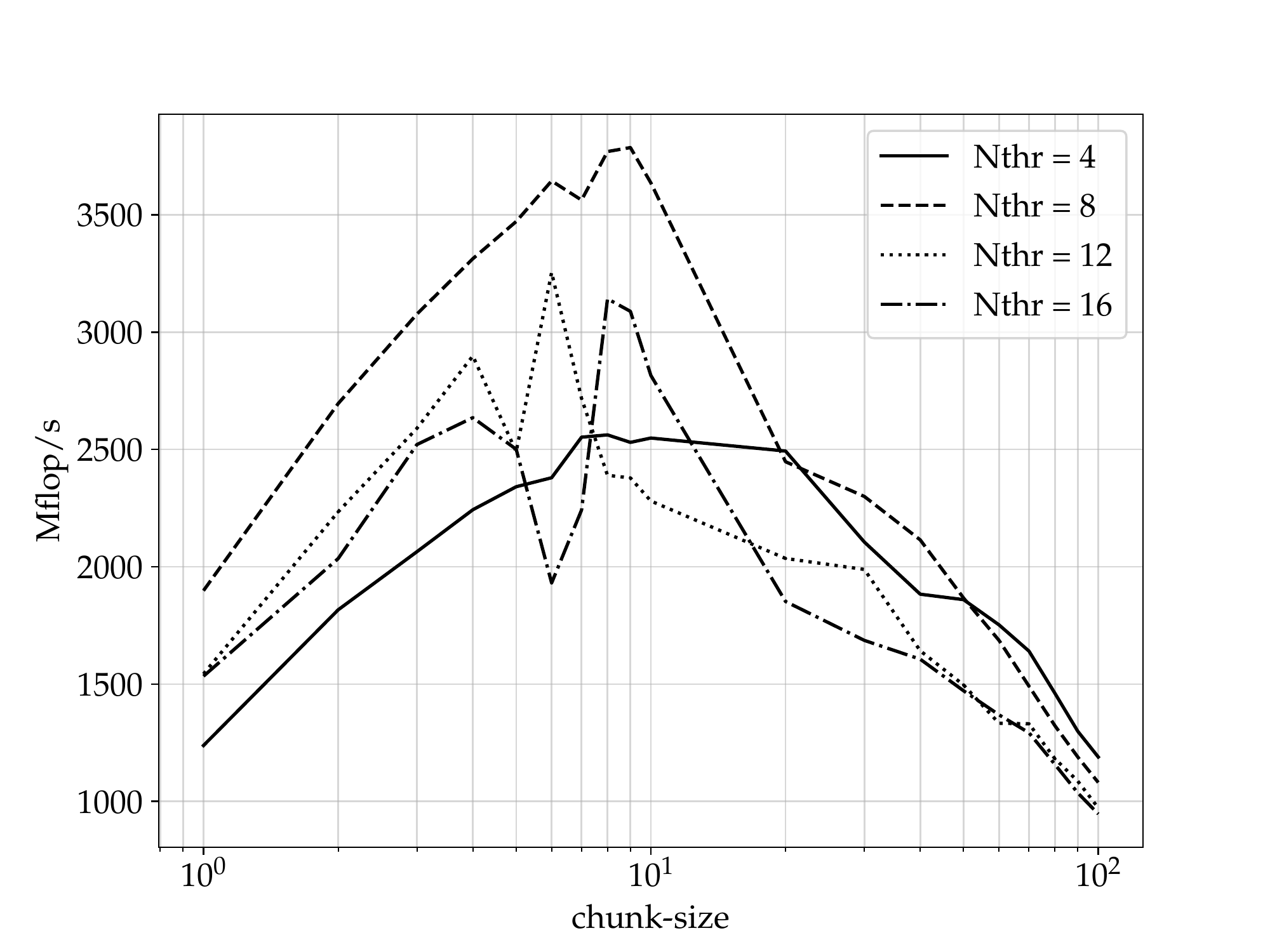}
\caption{Performance with respect to chunk-size for dense matrix dense matrix addition (dmatdmatadd) with ms=500 and 125 iterations  and with 4, 8, 12, 16 threads}
\label{fig:perf3}
\end{minipage}\hfill
\begin{minipage}[t]{\dimexpr.5\textwidth-1em}
  \centering
\includegraphics[ height=5cm, width=8cm]{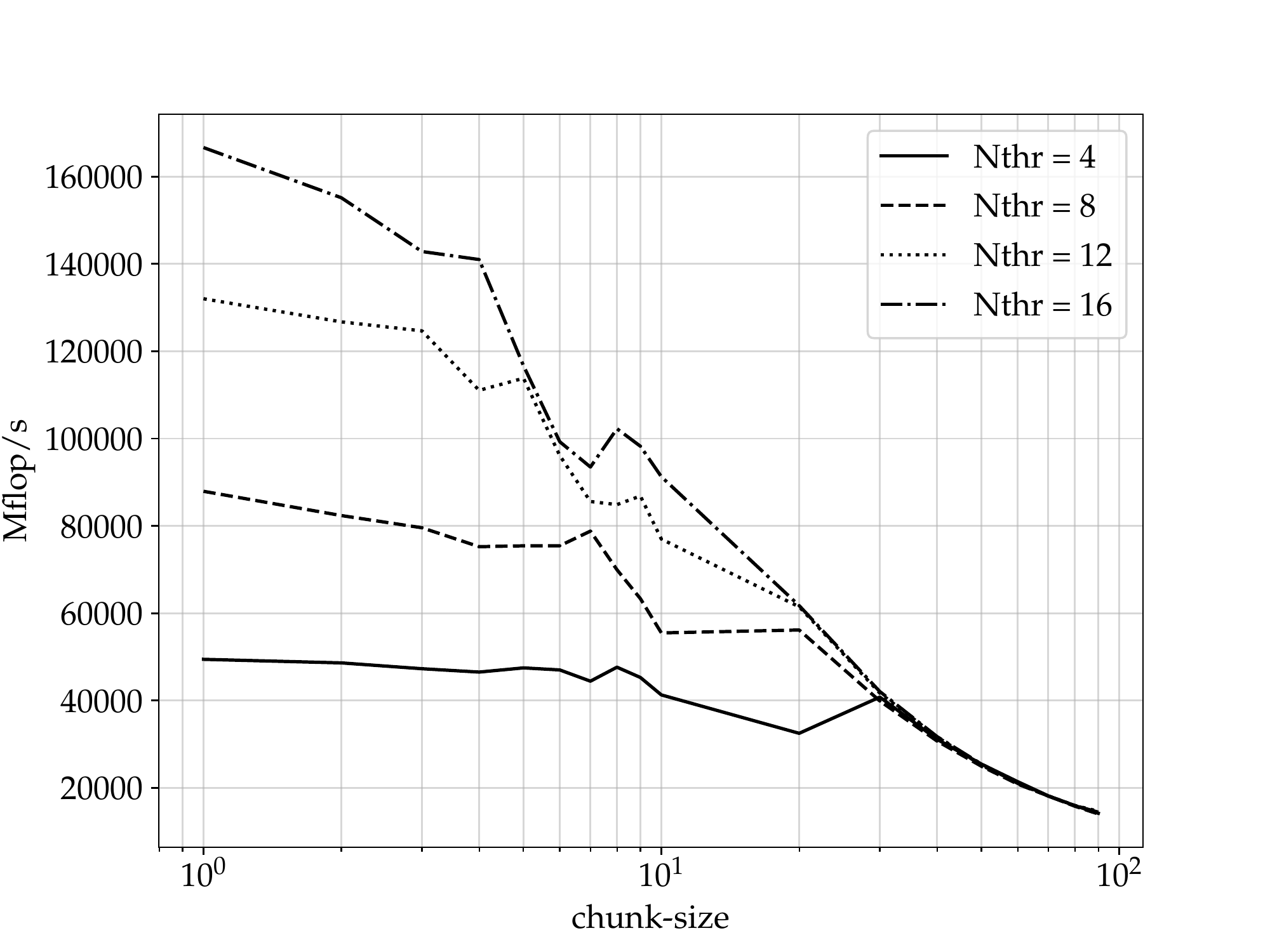}
\caption{Performance with respect to chunk-size for dense matrix dense matrix multiplication (dmattdmatmult) with ms=2500 and 100 iterations  and with 4, 8, 12, 16 threads}
\label{fig:perf4}
\end{minipage}
\end{figure}

In Figures~\ref{fig:perf1}--\ref{fig:perf3}, we see the expected bell-shape for performance. When chunk-sizes too small were chosen we observed high scheduling overheads and when chunk-sizes were too large, we observed poor parallelism. We note that, in some cases, there is a very specific chunk-size that yields a peak in performance. In others, there is a range of values of chunk-size that yield an optimal performance. For these cases, predicting the exact value of the optimal chunk-size is not necessary. In Figure~\ref{fig:perf1}, the performance line collected with 16 threads is a good example where all miss-classifications of chunk-sizes do not have the same impact on performance. For that specific benchmark and number of threads, $\var{cs}^*=9$. However, miss-classifying it with $\hat{\var{cs}}^*=8$ leads to a 14\% loss in performance and  miss-classifying it with $\hat{\var{cs}}^*=10$ causes a 40\% loss in performance. This is a concrete example that shows why we decided to build a custom classification model to predict chunk-sizes.
In figure~\ref{fig:perf4}, we see that the optimal chunk-size is always one. 
This could be due to the fact that, since matrix multiplications are very complex algorithms, each iteration of the \lstinline[mathescape]{for$\_$loop()}
involves significant work in comparison to the HPX's thread scheduling overhead. 
However, we suspect that this behavior is mainly caused by the specific kernels that Blaze calls. 
Taken together, the graphs indicate that the optimal chunk-size is always between one and ten so we used this insight to fix the set $\var{CS}$ to $\{1, 2, 3, ..., 9, 10\}$. The time to generate our data set of 288 examples was around two days which we view as reasonable so we did not attempt to reduce the size of the set $\var{CS}$. Throughout the rest of the paper, machine learning models will be limited to analyze and predict chunk-sizes from this set.

\subsection{Model Selection and True Error estimation}
\label{seq:results:model_selection}
As explained earlier, k-fold cross validation was used to select the best black box model. The black box models were also compared to two reference models; random choice and equal share model $\hat{\var{cs}}^* = \text{min}(10, \lceil\sfrac{N_{\var{ite}}}{N_{\var{thr}}}\rceil)$. Random choice is a reference point for the worst classifier one could achieve so it gives a lower bound on the error. The equal share model is a more naive model that simply tries to have one chunk per thread. We have previously stated that our models should only output values within the set $\var{CS}$ which is why, in the equal share model, an output larger than 10 is reduced to 10. This constraint is important if we are to use the MSOP to compare all the models.

\begin{table}[tbp]
\begin{center}
\resizebox{8cm}{1.6cm}{
\begin{tabular}{c|c|c}
\toprule
\textbf{Model}   & \textbf{MSOP} (\%) & \textbf{Prediction Time} ($\upmu$s) \\ \midrule
PreTO  - customDTC  & $94 \pm 1$        & $129 \pm 12$                                       \\ \hline
PreTO - DTC      & $94 \pm 2$        & $206 \pm 53$                                     \\ \hline
PreTO - LogReg   & $80 \pm 10$         & $251 \pm 254$                                      \\ \hline
PosTO - MLP      & $88 \pm 6$        & $1386 \pm 256$                                     \\ \hline
PosTO - DTR      & $93 \pm 2$        & $518 \pm 379$                                      \\ \hline
Random Guess  & $83 \pm 2$        & -                                               \\ \hline
Equal Share    & $88 \pm 2$        & -                                               \\ \bottomrule
\end{tabular}
}
\end{center}
\caption{Model Validation. The MSOP and prediction times are computed using k-fold cross-validation with k=3 on 192 examples. MSOP gives an idea of how relatively far, on average, the performance is from the optimal.}
\label{tab:tab_assess}
\end{table}

Table ~\ref{tab:tab_assess} provides the results from all models described above. We can see that all models except PreTO - LogReg, are significantly better than Random Guessing. Also, we see that the three models with the largest MSOP use an decision tree model, which suggest that trees are better for this specific task. As expected, PosTO models have a bigger average prediction time than PreTO models. Finally, we see that the custom tree has the best MSOP and the smallest average prediction time. 
Since the custom decision tree and scikit-learn decision trees do not have the same implementation, this difference in prediction time cannot be attributed solely to the complexity of the tree. We have therefore computed the average number of nodes for both our custom decision tree and scikit-learn decision tree classifier as a way to compare complexity. The average number of nodes for the custom and classical decision tree were $73 \pm 8$ and $ 155 \pm 14$ respectively. This shows that using the MSOP to train a classification tree allows to get good predictions of chunk-size with half the complexity of a classical decision tree. Since PreTO - customDTC has the best MSOP and minimal prediction time, it was selected as the model to be implemented in Blaze. After selecting the model, its overall performance on the test set of 96 examples was estimated with an MSOP of 94.8 (\%).

This shows that the ProTO model with an intrinsic custom Decision Tree Classifier can be generalized to examples outside of the training data. Since the training and test sets were obtained by shuffling all 288 examples for 6 benchmarks, we cannot conclude that the model would perform well in the cases of completely new benchmarks. The result simply states that if one restricts oneself to a specific architecture and a limited set of linear algebra benchmarks, one can generate training data on these benchmarks by using random values of $\var{ms}$ and $N_{\var{thr}}$. After training, this model can be used to get predictions on the same set of  benchmarks for all possible $\var{ms}$ and $N_{\var{thr}}$.

\subsection{C++ Implementation and Performance}
\label{sec:results:Cpp_implementation}
The Custom Decision Tree generated with the training set was implemented in Python so that we could compare it directly to the scikit-learn library. However, to get the predictions within Blaze, the tree needed to be implemented in C++. In order to do so, a function called \lstinline| printTreeHeader()| was written in Python to read the structure of the tree and output a header file that contained the tree structure.

\begin{lstlisting}[frame=single, language=C++, caption=Very Basic Example of Tree Structure transferred from a Python object to a C++ header File, float]
// Tree Structure in C++ Generated via Python Script
template <typename T>
inline int decisionTree(const std::vector<T>& featureVector) { 
    if (featureVector[1] < 0.4) {
        return 2;
    } 
    else {
        if (featureVector[0] < 1.0) {
            return 2        
        }
        else {
            return 1;
        } 
    }
}

\end{lstlisting}

The features are extracted via the functions described in Section ~\ref{seq:learning:model_selection} and the blocking described in Table ~\ref{tab:block} was also implemented in Blaze. The implementation in Blaze was designed to be as seamless as possible. To use the new machine learning scheduler, a user is simply required to add a flag when compiling.
By implementing the machine learning model in Blaze, we are able to assess how well it works in a real-life scenario. However, the relative measure provided by MSOP fails to describe the absolute performance of the models. Because of that, MSOP does not tell how the machine learning based scheduling compares to the one used in the previous implementation of the HPX backend of Blaze. The previous implementation is described in the last paragraph of Section~\ref{sec:metho:math}. To compare with the previous scheduling and blocking, we ran all benchmarks from Table~\ref{tab:allben} using the old backend, the new backend with equal sharing, that is $\hat{\var{cs}}^* = \lceil\sfrac{N_{\var{ite}}}{N_{\var{thr}}}\rceil$, and the new backend with chunk-size predicted via machine learning. The old backend and the new backend with equal sharing have approximately the same amount of work in a chunk. The only difference is that in the new backend, a chunk contains many blocks that are designed to work well with the cache. We, therefore, expect the difference between these two methods to only be caused by blocking. The difference in performance between the new backend with equal share and the new backend with machine learning scheduling is caused by the selection of chunk-size, since the block-sizes used are the same.

\begin{figure*}[tbp]
\centerline{\includegraphics[ scale=0.45]{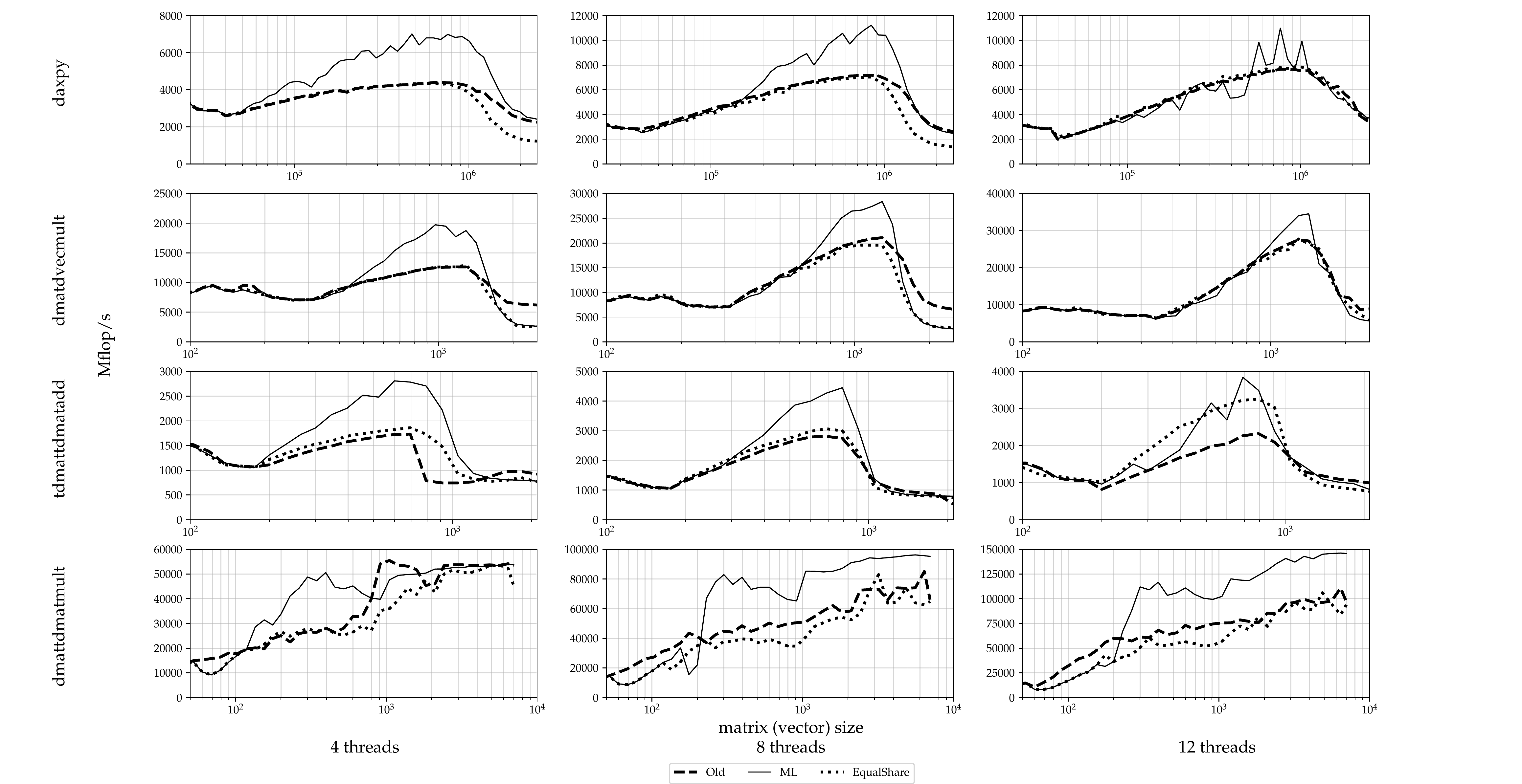}}
\caption{We show a subset of the benchmarks in Table ~\ref{tab:allben};  Daxpy, DMatDVecMult, TDMatTDMatAdd, DMatTDMatMult (top row to bottom row) with the old backend (dashed line), the new backend with the machine learning scheduler (full line) and the new backend with the equal share model for chunk size (dotted line). Differences between dashed and dotted lines can be attributed to blocking and changes from dotted to full are attributed to the machine learning predictions of chunk-size.}
\label{fig:MLvsOld}
\end{figure*}

Figure~\ref{fig:MLvsOld} shows a comparison of the three parallelization schemes for four different benchmarks from Table~\ref{tab:allben} (Daxpy, DMatDVecMult, TDMatTDMatAdd, DMatTDMatMult). 
Comparing the dotted and full lines shows the effect of the machine learning based chunking. Comparing the dashed and dotted lines shows the effect of blocking on performance. Since our main focus is to predict chunk-size, we are mainly interested in analyzing variations of performance that are attributed to the chunk-size prediction. Because of that, we will not attempt to analyze differences of performance attributed to blocking. The dotted line will simply be used as a reference point to show which variations in performance are due to blocking and which are due to chunk-size prediction. By looking at the Daxpy benchmark (first row of the graph), we see that the chunk-size selection improves performance for vector sizes ranging from $10^5$ to $10^6$. However, for 16 threads, the chunk-size predicted by machine learning yields very similar performance to the old backend implementation. We also observe more oscillations in performance for the machine learning scheduler. This can be attributed to the fact that there is more variance in the scheduling when more threads are used, thus, the machine learning struggles to predict the correct chunk-size. For DMatDVecMult, (second row), we observe an improvement in performance based on the chunk-size selection for vectors of size ranging from 500 to 1050. However, the performance for the machine learning scheduler drops for vector sizes beyond 1050. This drop seems to be related to blocking, not the machine learning model, since the performance also drops for the dotted lines.
By analyzing TDMatTDMatAdd (third row), we observe an improvement for matrix sizes in the range 200 to 1000. The most surprising result is that for 16 threads, the machine learning scheduler performs worst than the equal share model. Once again, the model seems to struggle when working on 16 threads.
An analysis of DMatTDMatMult (fourth row) shows that the machine learning scheduler yields a better performance for matrices of sizes ranging from 110 to $10^4$. The machine learning learning scheduler is better than equal share because, as shown in Figure ~\ref{fig:perf4}, the optimal chunk-size is always one, which the machine learning model is able to predict. For matrices of size smaller than 100, the performance is worst than the old backend. This is because the block-size used for these matrices is $(64, 64)$
which is so big that it can generate less blocks than there are threads. A fine-tuning of block-size in these cases should solve the issue.

\section{Conclusions and Perspectives}
\label{sec:conclusion}

\subsection{Conclusions}
In conclusion, our results show that, when using a certain blocking on our given architecture, we are able to calculate an optimal chunk-size for parallel linear algebra algorithms using machine learning models. We were also able to compare the model that predict the chunk-size directly (PreTO) and the model that predicts the performance to guide the choice of chunk-size (PosTO). In the literature, these two approaches were always used in isolation. Reasonable predictions of chunk-size were achieved by both models but the PosTO had larger prediction times due to the optimization step required at run-time. Also, we developed our own custom decision tree model using the MSOP as loss-function. This resulted in a decision tree that yields similar performance to classical decision trees but with half the number of nodes. The custom decision tree was seamlessly implemented in Blaze, only requiring a compile-time flag to launch the machine learning based scheduler. Early results of this scheduler show an improvement in performance when compared to the previous implementation of the HPX's backend. Nevertheless, we observed that the decision tree would struggle as the number of threads increased.

\subsection{Future Works}
One limitation of the proposed method is that it was only tested on one type of architecture. For the methodology to be seen as viable, it must be portable for multiple architectures. Two critical experiments must be done in the future to show such viability. We must demonstrate that the methodology can be reproduced on another architecture and we must show that a single model can be trained using training data collected from multiple architectures. This will require the use of architecture features in the feature vector. Moreover, a more efficient blocking method should be developed to avoid the cost of empirically computing the best block-size, This burden on the user inhibits the benefits of our approach. Other less critical research directions include: finding new features to characterize an operation type, include linear algebra operations with sparse matrices, and adding performance counters to the feature vector. Using performance counters will significantly change the work-flow of the Blaze user but there may be something useful to learn from such experiment. To conclude, we believe that this work lays a good foundation for more research that can be done in the domain of machine learning applied to parallel linear algebra. However, a lot more work is required to assess the viability of the proposed methodology in real-life applications.

\section*{Supplementary material}
All the artifacts of the paper are available on github\footnote{\url{https://github.com/gablabc/BlazeML}} and on Zenodo~\cite{doi_data}. The modified verison of Blaze is available here\footnote{\url{https://bitbucket.org/gablabc/blaze/src/hpx_backend_ML_Blaze3.4/}}.

\section*{Acknowledgment}

This work is funded by the Department of Energy Award DE-NA0003525. Any opinions, findings, conclusions or recommendations expressed in this material are those of the authors and do not necessarily reflect the views of the Department of Energy.
Special thanks to Google Summer of Code 2018 for supporting the first three months of the research.
Special thanks to Marc Laforest for his feedback on the paper.

\bibliographystyle{unsrt}  
\bibliography{references}

\end{document}